\documentclass[runningheads]{llncs}
\usepackage{amsmath} 
\usepackage{amsxtra} 
\usepackage{amssymb} 
\usepackage{graphicx}
\usepackage{mathtools}
\usepackage{multirow}
\usepackage[]{algorithm}
\usepackage{algorithmic}
\usepackage{comment}

\DeclareMathOperator*{\argmin}{arg\,min}

\newlength\myindent
\begin{document}
\title{RL-Based Method for Benchmarking the Adversarial Resilience and Robustness of Deep Reinforcement Learning Policies}
\titlerunning{Benchmarking Resilience and Robustness in DRL}
%
\author{Vahid Behzadan\inst{1} \and
William Hsu\inst{1}\\
\authorrunning{V. Behzadan et al.}
%
\institute{Kansas State University}
\email{\{behzadan, bhsu\}@ksu.edu}}
%
\maketitle              
\begin{abstract}
 This paper investigates the resilience and robustness of Deep Reinforcement Learning (DRL) policies to adversarial perturbations in the state space. Accordingly, we first present an approach for the disentanglement of vulnerabilities caused by representation learning of DRL agents from those that stem from the sensitivity of the DRL policies to distributional shifts in state transitions. Building on this approach, we propose two RL-based techniques for quantitative benchmarking of adversarial resilience and robustness in DRL policies against perturbations of state transitions. We demonstrate the feasibility of our proposals through experimental evaluation of resilience and robustness in DQN, A2C, and PPO2 policies trained in the Cartpole environment.

\end{abstract}
\section{Introduction}
Since the reports by Behzadan \& Munir \cite{behzadan2017vulnerability} and Huang et al. \cite{huang2017adversarial}, the primary emphasis of the state of the art in DRL security \cite{behzadan2018faults} has been on the vulnerability of policies to state-space perturbations. In particular, the manipulation of the policy via adversarial examples \cite{goodfellow6572explaining} has remained the main focus of current literature on this issue. However, this bias towards adversarial example attacks gives rise to a critical shortcoming: the analyses of such attacks fail to disentangle the vulnerability caused by the learned representation and that which is due to the sensitivity of the DRL dynamics to distributional shifts in state transitions.  Also, the performance of defenses proposed for adversarial example attacks are inherently limited to the considered attack mechanisms. As the most successful technique for mitigation of adversarial examples, adversarial training is known to enhance the robustness of machine learning models to the type of attack used for generating the training adversarial examples, while leaving the model vulnerable to other types of attacks\cite{tramer2017ensemble}. Furthermore, the current literature fails to provide solutions and approaches which can be used in practice to evaluate and improve the robustness and resilience of DRL policies to attacks that exploit the sensitivity to state transitions. Also, there remains a need for quantitative approaches to measure and benchmark the resilience and robustness of DRL policies in a reusable and generalizable manner.

In response to these shortcomings, this paper aims to address the problem of quantifying and benchmarking the robustness and resilience of a DRL agent to adversarial perturbations of state transitions at test-time, in a manner that is independent of the attack type. This improves the generalization of current techniques that analyze the model against specific adversarial example attacks. Accordingly, the main contributions of this paper are as follows:

\begin{enumerate}
    \item We present formulations of the resilience and robustness problems that enable the disentanglement of limitation in representation learning from sensitivity of policies to state transition dynamics.
    \item We propose two RL-based techniques and corresponding metrics for the measurement and benchmarking of resilience and robustness of DRL policies to perturbations of state transitions, 
    \item We demonstrate the feasibility of our proposal through experimental evaluation of their performance on DQN, A2C, and PPO2 agents trained in the Cartpole environment.
\end{enumerate}

The remainder of this paper is organized as follows: Section \ref{sec:Problem} defines and formulates the problems of adversarial resilience and robustness in DRL. Our proposed methods for benchmarking the test-time resilience and robustness of DRL policies are presented in Sections \ref{sec:TestResilience} and \ref{sec:TestRobustness}. Section \ref{sec:Setup} provides the details of experimental setup for evaluating the performance of our proposals, with the corresponding results presented in Section \ref{sec:Results}. The paper concludes in Section \ref{sec:Conclusion} with a summary of findings and remarks on future directions of research. 

\section{Problem Formulation}
\label{sec:Problem}
We consider the the generic problem of RL in the settings of a Markov Decision Process (MDP), described by the tuple $MDP := <\mathbb{S}, \mathbb{A}, \mathbb{R}, \mathbb{P}>$, where $\mathbb{S}$ is the set of reachable states in the process, $\mathbb{A}$ is the set of available actions, $\mathbb{R}$ is the mapping of transitions to the immediate reward, and $\mathbb{P}$ represents the transition probabilities (i.e., state dynamics), which are initially unknown to RL agents. At any given time-step $t$, the MDP is at a state $s_t\in \mathbb{S}$. The RL agent's choice of action at time $t$, $a_t \in \mathbb{A}$ causes a transition from $s_t$ to a state $s_{t+1}$ according to the transition probability $P(s_{t+1}|s_t , a_t)$. The agent receives a reward $r_{t+1} = R(s_t, a_t, s_{t+1})$ for choosing the action $a_t$ at state $s_t$. Interactions of the agent with MDP are determined by the policy $\pi$. When such interactions are deterministic, the policy $\pi: S\rightarrow \mathbb{A}$ is a mapping between the states and their corresponding actions. A stochastic policy $\pi(s)$ represents the probability distribution of implementing any action $a\in \mathbb{A}$ at state $s$. The goal of RL is to learn a policy that maximizes the expected discounted return $E[R_t]$, where $R_t = \sum_{k=0}^{\infty}\gamma^k r_{t+k}$; with $r_t$ denoting the instantaneous reward received at time $t$, and $\gamma$ is a discount factor $\gamma\in [0,1]$. 

To facilitate the formal statement of adversarial resilience and robustness, we first introduce the following definitions:
\begin{itemize}
    \item \textbf{\emph{Adversarial Regret}} at time $T$ is the difference between return obtained by the nominal (unperturbed) agent at time $T$ and the return obtained by the perturbed agent at time $T$. Formally: $\hat{R}_{adv}(T) = R_{nominal}(T) - R_{perturbed}(T)$. The time $T$ may represent either the terminal timestep of an episode, or the time-horizon of interest in the analysis. 
    \item \textbf{\emph{Adversarial Budget}} is defined by the one or more of the following parameters: the maximum number of features that can be perturbed in the observations ($O_{max}\in [0, \infty]$ ), the maximum number of observations that can be perturbed ( $N_{max}\in [0, \infty]$ ), and the probability of perturbing each observation ( $P(perturb)\in [0, 1]$ ). 
\end{itemize}

Building on these two concepts, we define the problems of adversarial resilience and robustness as follows:
\begin{enumerate}
    \item \textbf{\emph{Test-Time Resilience:}} The minimum number of state perturbations required to incur the maximum reduction to the total return at time $T$ (denoted by $\hat{R}_{adv}(T)$) for an agent driven by a policy $\pi(s)$ in an environment with transition dynamics $\mathbb{P}$.
    \item \textbf{\emph{Test-Time Robustness:}} The maximum adversarial regret $\hat{R}_{adv}(T) = \epsilon_{max}$ achievable via a maximum of $\delta_{max}$ state perturbations for an agent driven by a policy $\pi(s)$ in an environment with transition dynamics $\mathbb{P}$.
\end{enumerate}    

The following sections provide the details of our proposed solutions to each of the aforementioned problem settings.

\section{Benchmarking of Test-Time Resilience}
\label{sec:TestResilience}
This problem can be modeled as that of finding an optimal adversarial policy $\pi_{adv}(s)$ that minimizes the cost incurred to the adversary $C_{adv}$ in order to impose the maximum adversarial regret $\hat{R}_{adv}(T)$, the worst-case value of which is the highest cumulative reward achieved by the target policy $R_{max}$. Our proposed approach is through the formulation of this problem in the settings of reinforcement learning. The state space in the corresponding MDP is the set of states in the target MDP, augmented with the action of the target in that state, i.e., $S' = \{\forall s\in \mathbb{S}: (s, \pi(s))\}$. For the purpose of measuring a lower-bound for the resilience, we consider the worst-case white-box adversary, which is able to impose targeted state perturbations with $100\%$ success rate, to induce any action within the permissible action-set of the target $\mathbb{A}$ which has the lowest $Q$-value at any state $s$ according to the target's optimal state-action value function $Q^*$. In this case, the set of permissible adversarial actions at any state $s$ is given by:
\begin{equation}
    A_{adv} (s) = \{\text{No Action}\} \cup \mathbb{A}\setminus{\pi^*(s)}
\end{equation}
Where $\mathbb{A}$ is the action set of the targeted agent, and $\pi: S\rightarrow A$ is the policy of the targeted agent. In the proposed approach, the adversarial reward value is determined via the procedure detailed in Algorithm \ref{alg:resilience}:
\begin{algorithm}[H] 
\caption{Reward Assignment of RL Agent for Measuring Adversarial Resilience} 
\label{alg:resilience} 
\begin{algorithmic} 
    \REQUIRE Target policy $\pi^*$, Perturbation cost function $c_{adv}(., .)$, Maximum achievable score $R_{max}$, Optimal state-action value function $Q^*(.,.)$, Current adversarial policy $\pi^{adv}$, Current state $s_t$, Current count of adversarial actions $AdvCount$, Current score $R_t$
    \STATE Set ToPerturb $\leftarrow \pi^{adv}(s_t)$
    \IF{ToPerturb is False}
        \STATE $a_t \leftarrow \pi^*(s_t)$
        \STATE ${Reward} \leftarrow 0$
    \ELSE
        \STATE $a'_t \leftarrow \argmin_a Q^*(s_t, a)$
        \STATE ${Reward} \leftarrow - c_{adv}(s_t, a'_t)$
    \ENDIF
    \IF{either $s_t$ or $s'_t$ is terminal}
        \STATE ${Reward} += (R_{max} - R_t)$
    \ENDIF
\end{algorithmic}
\end{algorithm}

where $c(s_t, a'_t)$ is the cost of imposing the state perturbation which induces the adversarial action $a'_t$ at state $s_t$. It is noteworthy that if the value of $c(s_t, a'_t)$ is invariant with respect to $a'_t$, the adversarial action set reduces to:
\begin{equation}
    A_{adv} (s) = \{\text{No Action}, \text{\emph{Induce}}\argmin_a Q(s, a)\} 
\end{equation}

To obtain the test-time resilience of policy $\pi^*$ to state perturbations, we propose the following procedure:
\begin{enumerate}
    \item If the state-action value function of the target $Q^*$ is not available (i.e., black-box testing), approximate $Q^*$ via policy imitation \cite{hussein2017imitation}. 
    \item Train the adversarial agent against the target following $\pi$ in its training environment, report the optimal adversarial return $R_{perturbed}^*$ and the maximum adversarial regret $R^*_{adv}(T)$. 
    \item Apply the adversarial policy against the target in $N$ episodes, record total cost $C_{adv}$ for each episode,
    \item Report the average of $C_{adv}$ over $N$ episodes as the mean test-time resilience of $\pi$ in the given environment.
\end{enumerate}
This procedure introduces 3 metrics for the quantification of test-time resilience: the optimal adversarial return $R^*_{perturbed}$ achieved in the training process of the adversarial policy, the maximum adversarial regret $R^*_{adv}(T)$ achieved during training, and the mean per-episode of the total cost $C_{adv}$. These metrics provide the means to benchmark and compare the test-time resilience of different policies trained to optimize the agent's performance in a given environment.

For the purpose of measuring resilience, we consider convergence to be reached if the average adversarial regret over 200 episodes remains constant. This definition relaxes the instabilities that may arise due to the configuration and architecture of the DRL training process. It is noteworthy that depending on the training algorithm and design parameters, this procedure is not guaranteed to converge to the global optimal. However, by reporting the number of iterations and configuration of random number generators with a constant seed, the reported results present a reproducible loose lower bound on the adversarial resilience of the target. Also, the trained adversarial policy can be used to test other policies for comparison of such lower-bounds under the same adversarial strategy.


\section{Benchmarking of Test-Time Robustness}
\label{sec:TestRobustness}
For this problem, we propose a modified version of the procedure developed for benchmarking the test-time resilience. Accordingly, the reward function is adjusted to account for the lack of a target $\epsilon$, as well as the addition of an adversarial budget constraint $\delta_{max}$. The reward measurement of this process is outlined in Algorithm \ref{alg:robustness}:

\begin{algorithm}[H] 
\caption{Reward Assignment of RL Agent for Measuring Adversarial Robustness} 
\label{alg:robustness} 
\begin{algorithmic} 
    \REQUIRE Maximum perturbation budget $\delta_{max}$, Perturbation cost function $c_{adv}(., .)$, Maximum achievable score $R_{max}$, Optimal state-action value function $Q^*(.,.)$, Current adversarial policy $\pi^{adv}$, Current state $s$, Current count of adversarial actions $AdvCount$, Current score $R_t$
    \STATE Set AdversarialAction $\leftarrow \pi^{adv}(s)$
    \IF{AdversarialAction is NoAction}
        \STATE ${Reward} \leftarrow 0$
    \ELSIF{$AdvCount \geq \delta_{max}$}
        \STATE ${Reward} \leftarrow -c_adv(s, AdversarialAction) \times \delta_{max} $
        \STATE $AdvCount += 1$
    \ELSE
        \STATE ${Reward} \leftarrow -c_adv(s, AdversarialAction)$
        \STATE $AdvCount += 1$
    \ENDIF
    \IF{$s$ is terminal}
        \STATE ${Reward} += 1.0 * (R_{max} - R_t)$
        \STATE $AdvCount \leftarrow 0$
    \ENDIF
\end{algorithmic}
\end{algorithm}

The proposed procedure for measuring the test-time robustness of a given DRL policy to adversarial state perturbations is as follows:
\begin{enumerate}
    \item If the state-action value function of the target $Q^*$ is not available (i.e., black-box testing settings), approximate $Q^*$ from the policy using imitation learning (e.g., \cite{hussein2017imitation}), 
    \item Train the adversarial agent against the target policy $\pi^*$ in its training environment, report the maximum adversarial regret $R_{adv}^*(T)$ for time $T$ achieved at adversarial optimality,
    \item Apply the adversarial policy against the target for $N$ episodes, record the adversarial regret at the end of each episode $R_{adv}(T)$,
    \item Report the average of $R_{adv}(T)$ over $N$ episodes as the mean per-episode test-time robustness of $\pi^*$ in the given environment.
\end{enumerate}


\section{Experiment Setup}
\label{sec:Setup}
\textbf{Environment and Target Policies:} To demonstrate the performance of the proposed procedures for benchmarking the test-time robustness and resilience in DRL policies, we present the analysis of the aforementioned measurements for policies trained in the CartPole environment in OpenAI Gym \cite{brockman2016openai}. The considered policies are chosen to represent the commonly-adopted state of the art method from each class of DRL algorithms. From value-iteration approaches, we consider DQN with prioritized replay. From policy gradient approaches, we consider PPO2. As for actor-critic methods, we investigate the A2C method. Table \ref{CartPole} presents the specifications of the CartPole environment, and Tables \ref{CartPoleDQN} -- \ref{CartPolePPO2} provide the parameter settings of each target policy.

\begin{table}[H]
\centering
\label{CartPole}
\caption{Specifications of the CartPole Environment}
\begin{tabular}{|l|l|}
\hline
Observation Space & \begin{tabular}[c]{@{}l@{}}Cart Position {[}-4.8, +4.8{]}\\ Cart Velocity {[}-inf, +inf{]}\\ Pole Angle {[}-24 deg, +24 deg{]}\\ Pole Velocity at Tip {[}-inf, +inf{]}\end{tabular} \\ \hline
Action Space      & \begin{tabular}[c]{@{}l@{}}0 : Push cart to the left\\ 1 : Push cart to the right\end{tabular}                                                                                      \\ \hline
Reward            & +1 for every step taken                                                                                                                                                             \\ \hline
Termination       & \begin{tabular}[c]{@{}l@{}}Pole Angle is more than 12 degrees\\ Cart Position is more than 2.4\\ Episode length is greater than 500\end{tabular}                                    \\ \hline
\end{tabular}
\end{table}

\begin{table}[H]
\centering
\label{CartPoleDQN}
\caption{Parameters of DQN Policy}
\begin{tabular}{|l|l|}
\hline
No. Timesteps               & $10^5$                \\ \hline
$\gamma$                    & $0.99$                \\ \hline
Learning Rate               & $10^{-3}$             \\ \hline
Replay Buffer Size          & 50000                 \\ \hline
First Learning Step         & 1000                  \\ \hline
Target Network Update Freq. & 500                   \\ \hline
Prioritized Replay          & True                  \\ \hline
Exploration                 & Parameter-Space Noise \\ \hline
Exploration Fraction        & 0.1                   \\ \hline
Final Exploration Prob.     & 0.02                  \\ \hline
Max. Total Reward           & 500                   \\ \hline
\end{tabular}
\end{table}

\begin{table}[H]
\centering
\label{CartPoleA2C}
\caption{Parameters of A2C Policy}
\begin{tabular}{|l|l|}
\hline
No. Timesteps              & $5\times 10^5$    \\ \hline
$\gamma$                   & $0.99$            \\ \hline
Learning Rate              & $7\times 10^{-4}$ \\ \hline
Entropy Coefficient        & 0.0               \\ \hline
Value Function Coefficient & 0.25              \\ \hline
Max. Total Reward          & 500               \\ \hline
\end{tabular}
\end{table}

\begin{table}[H]
\centering
\label{CartPolePPO2}
\caption{Parameters of A2C Policy}
\begin{tabular}{|l|l|}
\hline
No. Environments                    & 8                 \\ \hline
No. Timesteps                       & $10^6$            \\ \hline
No. Runs per Environment per Update & 2048              \\ \hline
No. Minibatches per update          & 32                \\ \hline
Bias-Variance Trade-Off Factor      & 0.95              \\ \hline
No. Surrogate Epochs                & 10                \\ \hline
$\gamma$                            & $0.99$            \\ \hline
Learning Rate                       & $3\times 10^{-4}$ \\ \hline
Entropy Coefficient                 & 0.0               \\ \hline
Value Function Coefficient          & 0.5               \\ \hline
Max. Total Reward                   & 500               \\ \hline
\end{tabular}
\end{table}

\textbf{Adversarial Agent:} In these experiments, the adversarial agent is a DQN agent with the hyperparameters provided in Table \ref{AdvDQN}. We consider a homogeneous perturbation cost function for all state perturbations, that is $\forall s, a': c_{adv}(s, a') = c_{adv}$. For both the resilience and robustness measurements, we set $c_{adv} = 1$ (i.e., each perturbation incurs a cost of $1$ to the adversary). The training process is terminated when the adversarial regret is maximized and the 100-episode average of the number of adversarial perturbations is quasi-stable for 200 episodes. 

\begin{table}[H]
\centering
\label{AdvDQN}
\caption{Parameters of DQN Policy}
\begin{tabular}{|l|l|}
\hline
Max. Timesteps              & $10^5$                \\ \hline
$\gamma$                    & $0.99$                \\ \hline
Learning Rate               & $10^{-3}$             \\ \hline
Replay Buffer Size          & 50000                 \\ \hline
First Learning Step         & 1000                  \\ \hline
Target Network Update Freq. & 500                   \\ \hline
Experience Selection        & Prioritized Replay    \\ \hline
Exploration                 & Parameter-Space Noise \\ \hline
Exploration Fraction        & 0.1                   \\ \hline
Final Exploration Prob.     & 0.02                  \\ \hline
\end{tabular}
\end{table}

\section{Results}
\label{sec:Results}
\subsection{Resilience Benchmarks}
We consider the white-box settings in the training of adversarial agents for resilience measurement. For the DQN target, the optimal state-action value function $Q^*$ of the target is directly utilized. As for the A2C and PPO2 targets, the state-action value function is calculated from the internally-available state value estimations $V*(s)$ according to the following transformation:
\begin{equation}
    Q^*(s_t, a) = r(s_t, a) + \gamma V^*(s_{t+1})
\end{equation}
where $s_{t+1}$ is the state resulting from a transition out of state $s_t$ by implementing action $a$.

\subsubsection{Training Results:} 
\begin{figure}[h]
	
	\centering
	
	\includegraphics[width = 0.5\columnwidth]{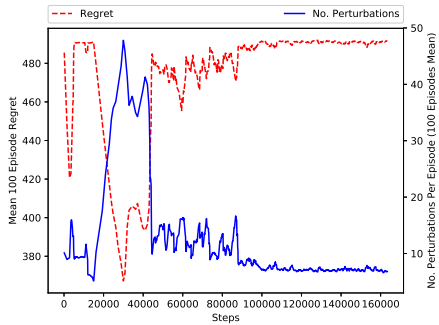}
	
	\caption{Adversarial Training Progress for Resilience Benchmarking of the DQN Policy}
	
	\label{fig:ResDQN}
	
\end{figure}

\begin{figure}[h]
	
	\centering
	
	\includegraphics[width = 0.5\columnwidth]{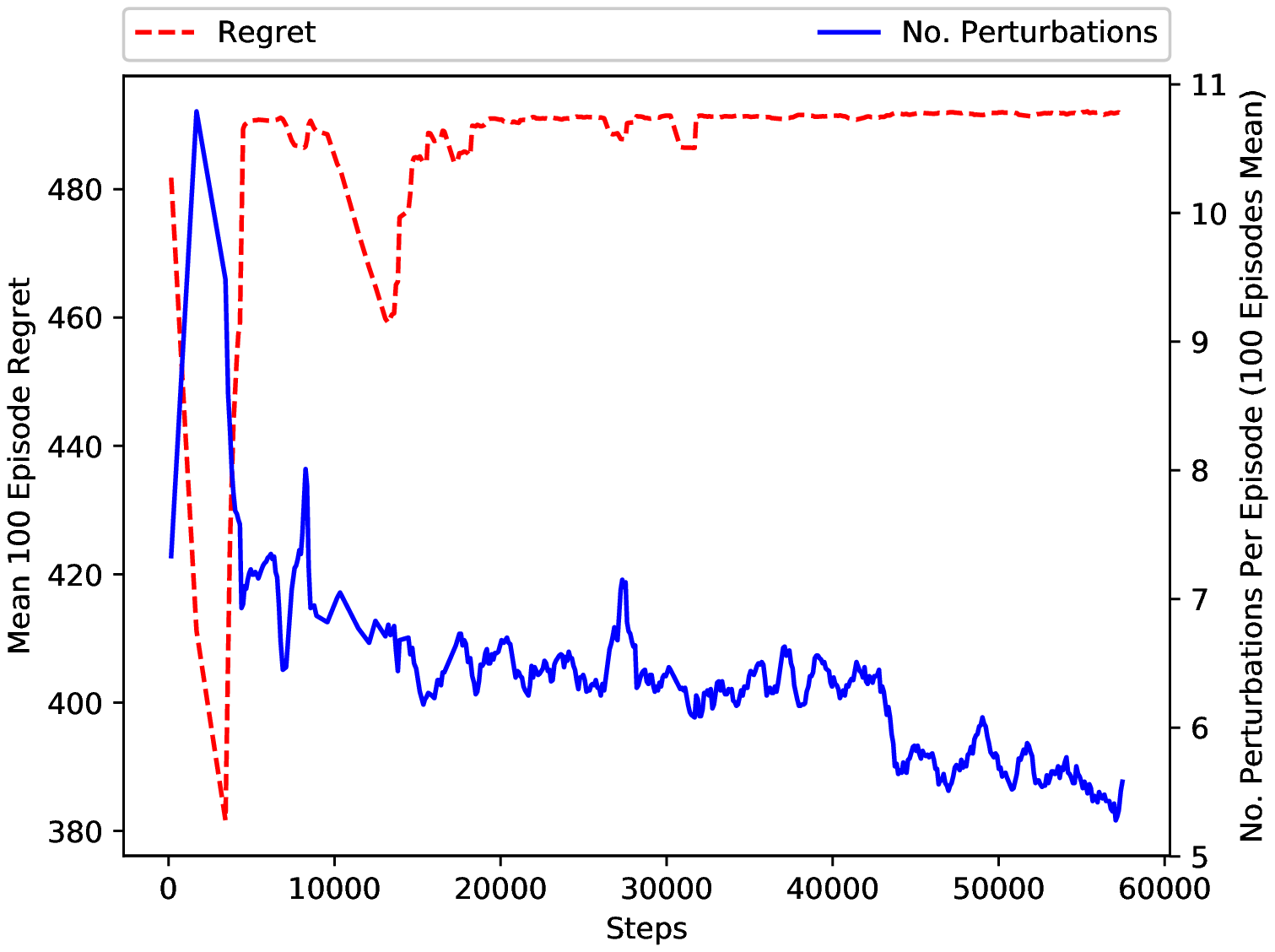}
	
	\caption{Adversarial Training Progress for Resilience Benchmarking of the A2C Policy}
	
	\label{fig:ResA2C}
	
\end{figure}

\begin{figure}[h]
	
	\centering
	
	\includegraphics[width = 0.5\columnwidth]{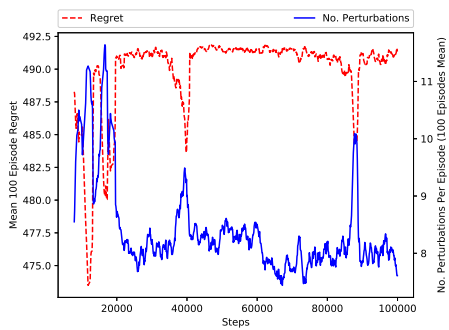}
	
	\caption{Adversarial Training Progress for Resilience Benchmarking of the PPO2 Policy}
	
	\label{fig:ResPPO}
	
\end{figure}

The training progress plots of adversarial DQN policy on the three target policies are presented in Fig.\ref{fig:ResDQN}--\ref{fig:ResPPO}. It can be seen that all three policies converge to the same optima. However, for the adversary targeting the DQN policies, the convergence is achieved at a higher number of training steps. 

It is noteworthy that for all three policies, the mean-per-100 episodes of the minimum number of perturbations at convergence is almost similar (as reported in Table \ref{Table:ResComparison}), with A2C having the largest value of $7.69$ perturbations, PPO2 at $7.49$ perturbations, and DQN having the lowest value of $7.13$. Also, the test-time performance of these trained policies indicate similar results, with DQN requiring $6.95$ perturbations to incur an adversarial regret of $491.15$, PPO2 requiring $7.72$ perturbations for an adversarial regret of $490.47$, and A2C requiring $8.71$ perturbations for an adversarial regret of $488.16$. Accordingly, we can interpret these results as follows: the DQN policy has the lowest adversarial resilience among the three, followed by the PPO2 policy. Within the context of this comparison, the A2C policy is found to be the most resilient to state-space perturbation attacks.

\subsection{Test-Time Step-Perturbation Distribution:}
\begin{figure}[h]
	
	\centering
	
	\includegraphics[width = 0.5\columnwidth]{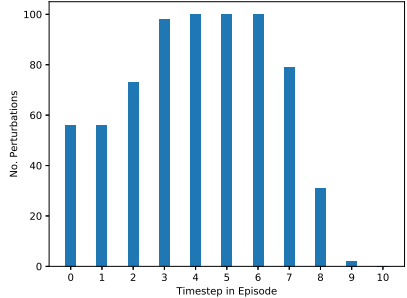}
	
	\caption{Perturbation Count Per Episodic TimeStep in 100 Runs Targeting DQN Policy}
	
	\label{fig:ResDQNTest}
	
\end{figure}

\begin{figure}[h]
	
	\centering
	
	\includegraphics[width = 0.5\columnwidth]{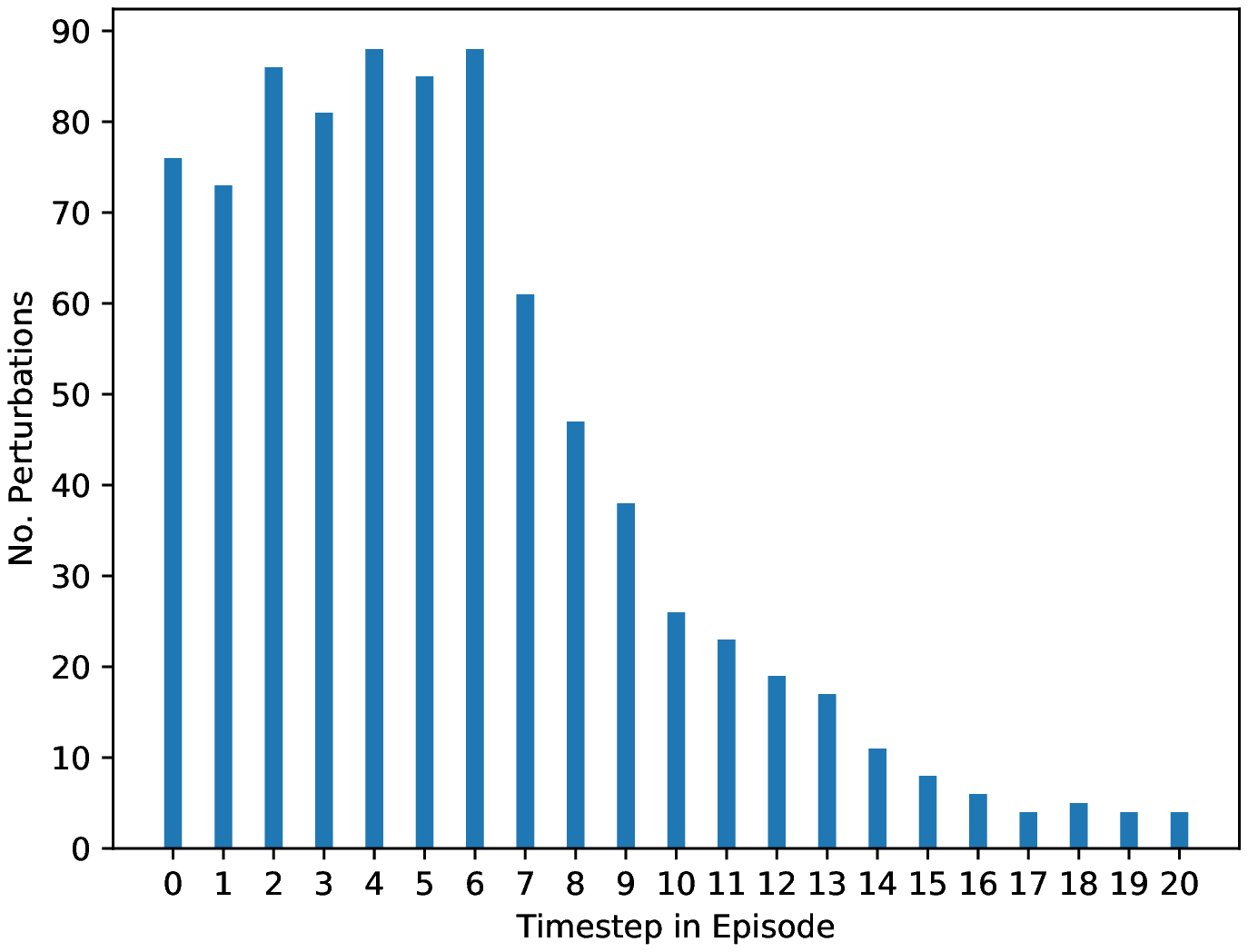}
	
	\caption{Perturbation Count Per Episodic TimeStep in 100 Runs Targeting A2C Policy}
	
	\label{fig:ResA2CTest}
	
\end{figure}

\begin{figure}[h]
	
	\centering
	
	\includegraphics[width = 0.5\columnwidth]{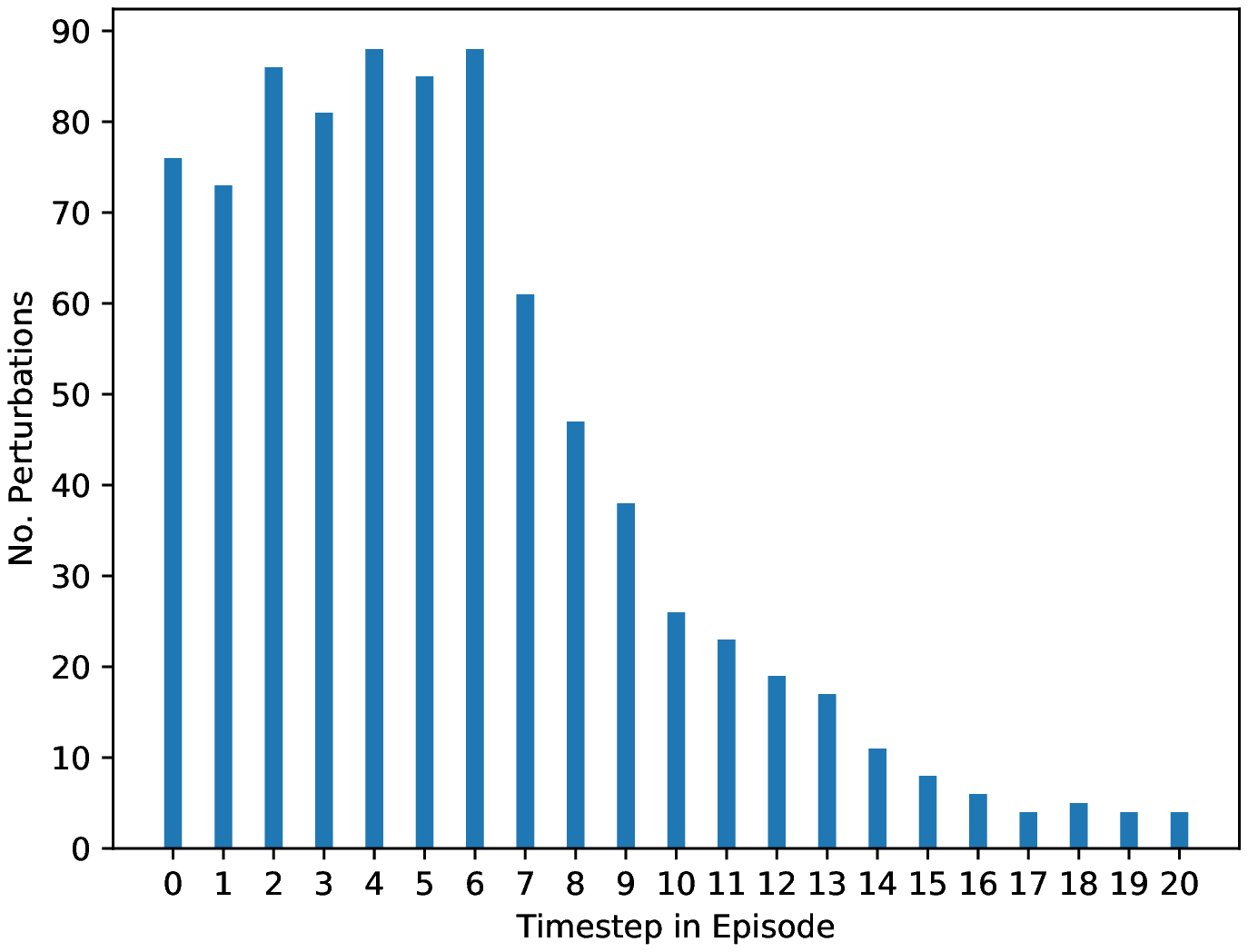}
	
	\caption{Perturbation Count Per Episodic TimeStep in 100 Runs Targeting PPO Policy}
	
	\label{fig:ResPPOTest}
	
\end{figure}

To investigate the state-transition vulnerability of each policy, we also study the frequency of perturbing states at each timestep of an episode for the three adversarial policies. The results, presented in Fig. \ref{fig:ResDQNTest} -- \ref{fig:ResPPOTest} illustrate that in all three policies, the initial timesteps have been the subject of most perturbations. This result is noteworthy, as it contradicts with the assumption of Lin et al.\cite{lin2017tactics} that the most effective adversarial perturbations are those that are mounted towards the terminal state of the environment. 

\begin{table}[H]
\label{Table:ResComparison}
\caption{Comparison of Test-Time and Training-Time Resilience Measurements for DQN, A2C, and PPO2 Policies}
\small
\resizebox{\textwidth}{!}{
\begin{tabular}{|c|c|c|c|c|c|}
\hline
\textbf{Target Policy} & \textbf{Max. Regret} & \multicolumn{1}{l|}{\textbf{Avg. Regret (Training)}} & \multicolumn{1}{l|}{\textbf{Avg. No. Perturbations (Training)}} & \multicolumn{1}{l|}{\textbf{Avg. Regret}} & \multicolumn{1}{l|}{\textbf{Avg. No. Perturbations}} \\ \hline
DQN                    & 492                  & 491.24                                               & 7.13                                                            & 491.15                                    & 6.95                                                 \\ \hline
A2C                    & 492                  & 491.44                                               & 7.69                                                            & 488.16                                    & 8.71                                                 \\ \hline
PPO2                   & 492                  & 491.72                                               & 7.49                                                            & 490.47                                    & 7.72                                                 \\ \hline
\end{tabular}}
\end{table}

\subsection{Robustness Benchmarks}
\begin{figure}[H]
	
	\centering
	
	\includegraphics[width = 0.4\columnwidth]{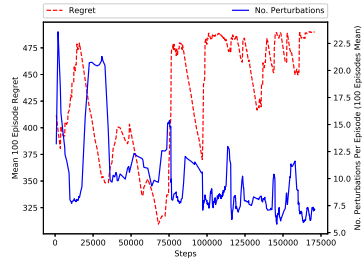}
	
	\caption{Adversarial Training Progress for Robustness Benchmarking - DQN, $\delta_{max} = 10$}
	
	\label{fig:RobustDQN}
	
\end{figure}

\begin{figure}[H]
	
	\centering
	
	\includegraphics[width = 0.4\columnwidth]{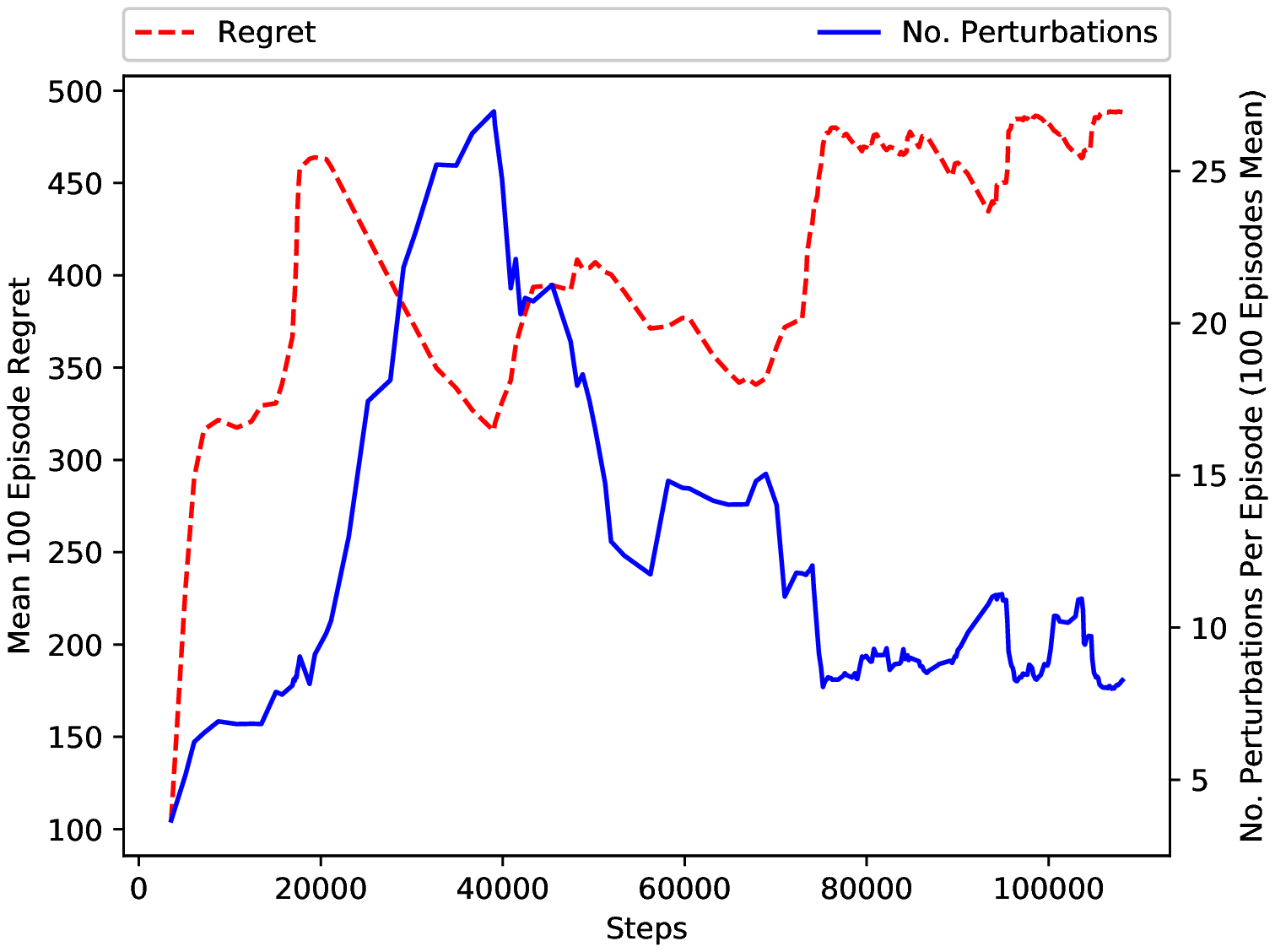}
	
	\caption{Adversarial Training Progress for Robustness Benchmarking - A2C, $\delta_{max} = 10$}
	
	\label{fig:RobustA2C}
	
\end{figure}

\begin{figure}[H]
	
	\centering
	
	\includegraphics[width = 0.4\columnwidth]{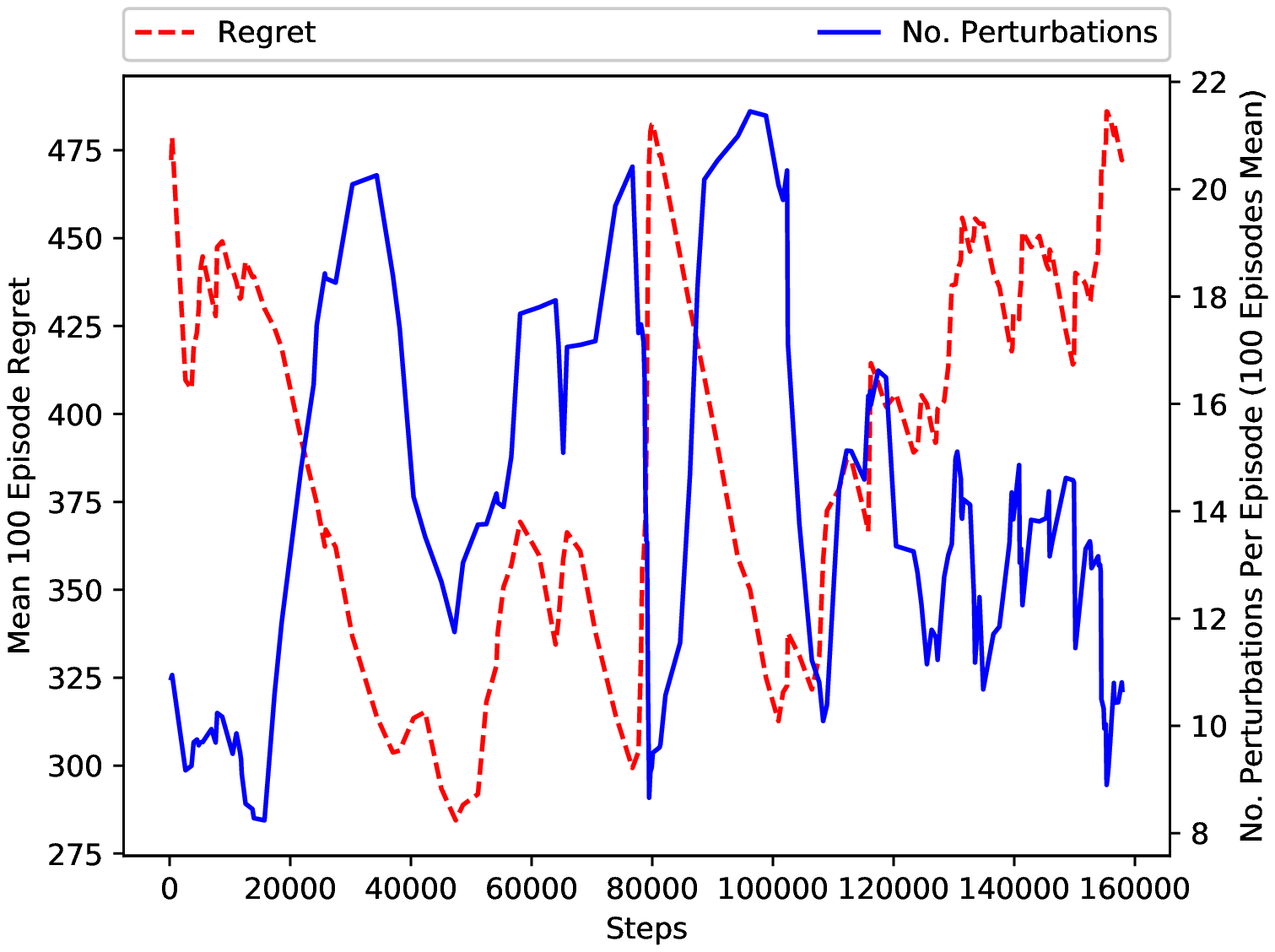}
	
	\caption{Adversarial Training Progress for Robustness Benchmarking - PPO2, $\delta_{max} = 10$}
	
	\label{fig:RobustPPO}
	
\end{figure}

\begin{figure}[H]
	
	\centering
	
	\includegraphics[width = 0.4\columnwidth]{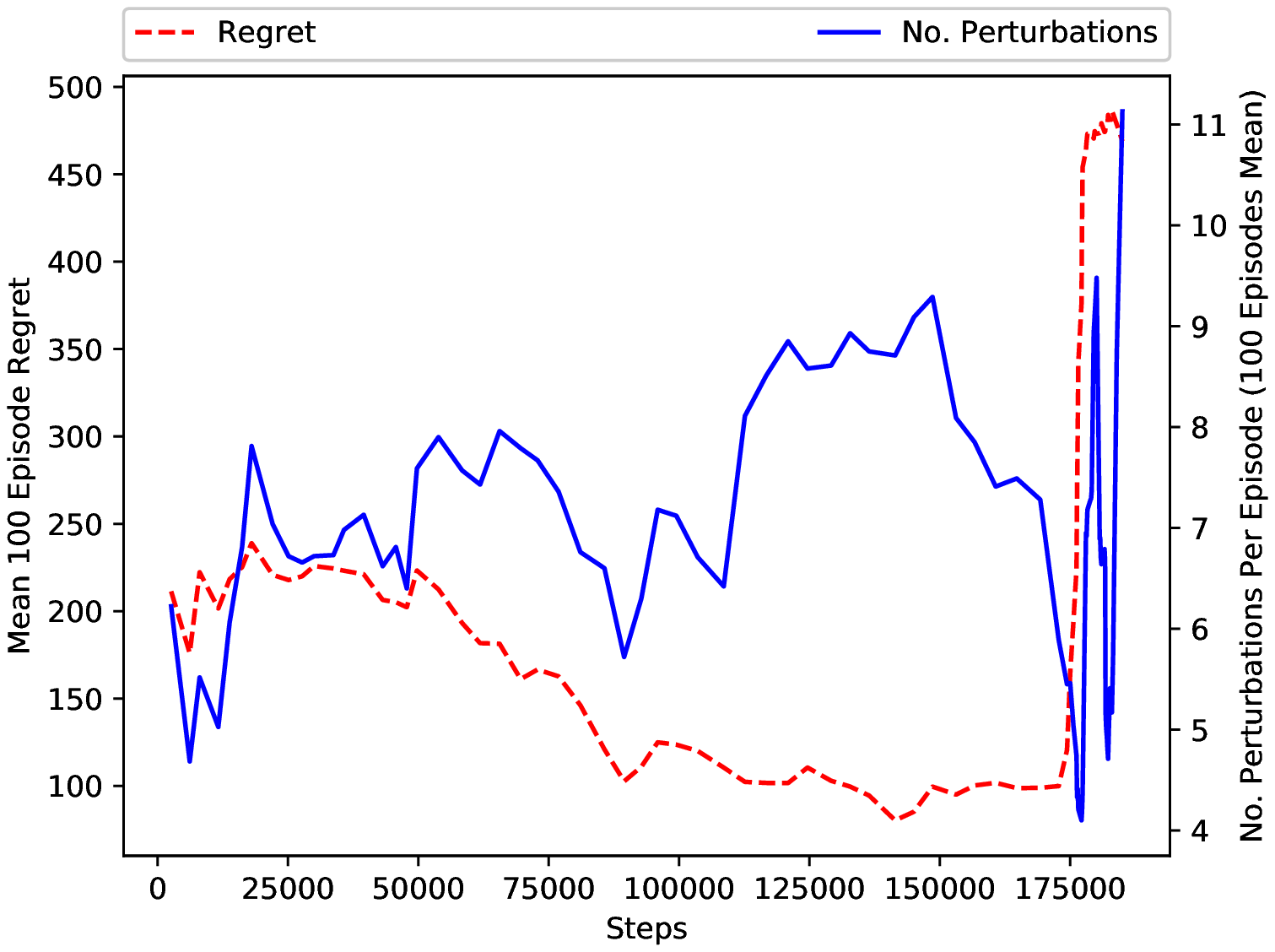}
	
	\caption{Adversarial Training Progress for Robustness Benchmarking - DQN, $\delta_{max} = 5$}
	
	\label{fig:Robust5DQN}
	
\end{figure}

\begin{figure}[H]
	
	\centering
	
	\includegraphics[width = 0.4\columnwidth]{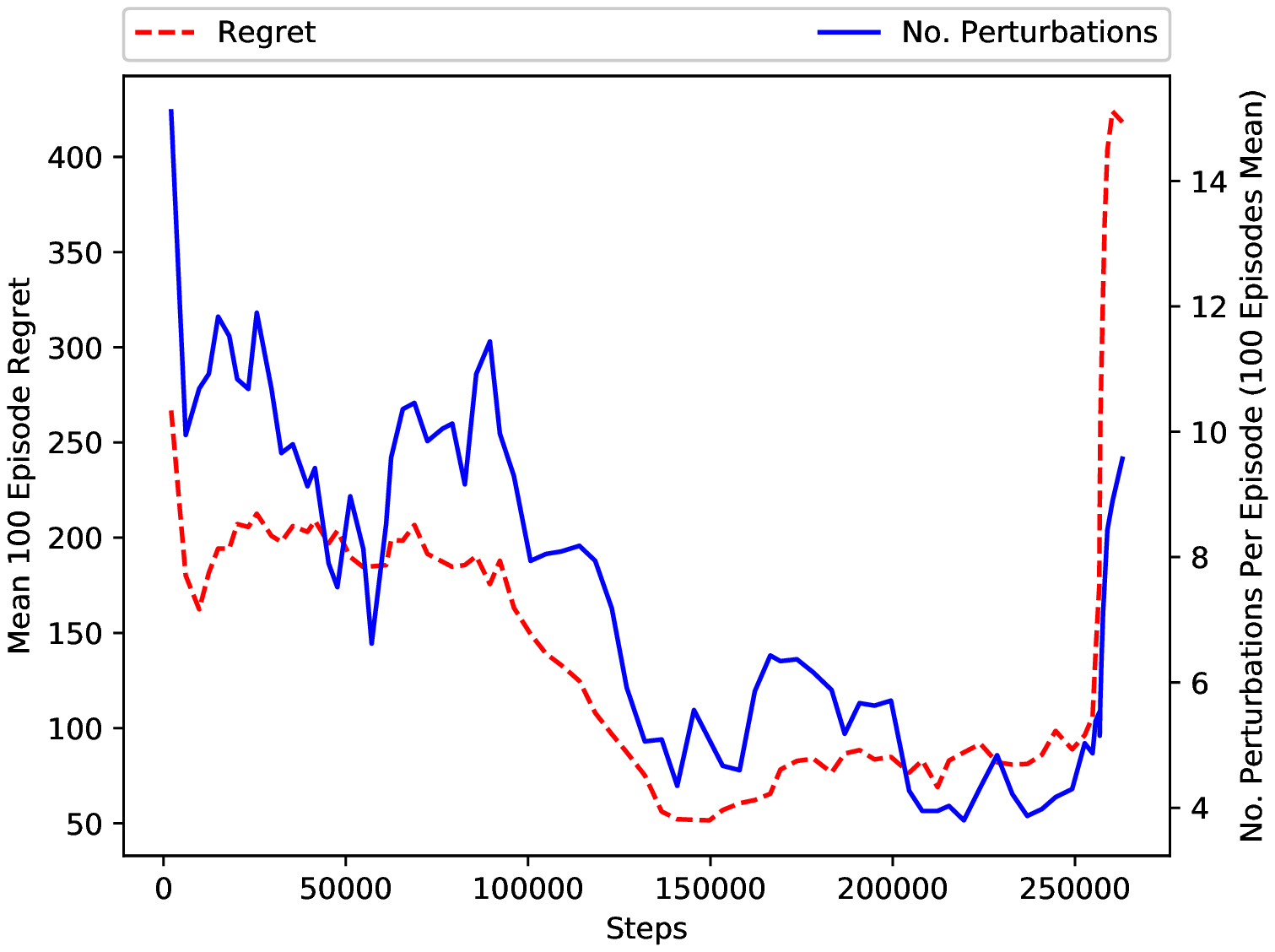}
	
	\caption{Adversarial Training Progress for Robustness Benchmarking - A2C, $\delta_{max} = 5$}
	
	\label{fig:Robust5A2C}
	
\end{figure}

\begin{figure}[H]
	
	\centering
	
	\includegraphics[width = 0.4\columnwidth]{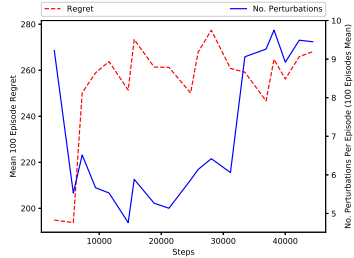}
	
	\caption{Adversarial Training Progress for Robustness Benchmarking - PPO2, $\delta_{max} = 5$}
	
	\label{fig:Robust5PPO}
	
\end{figure}

To demonstrate the performance of our proposed technique for benchmarking the robustness of DRL policies, we provide the training-time results for two cases of $\delta_{max} = 10$ and $\delta_{max} = 5$ for DQN, A2C, and PPO2 Policies. As illustrated in Fig.\ref{fig:RobustDQN} -- \ref{fig:RobustPPO}, all three adversarial policies converge with similar minimum perturbation counts as those obtained in resilience analysis. This is expected, as the resilience analysis established that the minimum number of actions required for maximum regret is $~7.5$, which is less than the available budget of $\delta_{max} = 10$ As for the case of $\delta_{max} = 5$, Fig.\ref{fig:Robust5DQN} -- \ref{fig:Robust5PPO} demonstrate significant differences between the three policies. In Fig.\ref{fig:Robust5DQN}, it can be seen that at 5 actions, the convergence occurs with an adversarial regret of $462.5$, while for A2C, the best 5-action indication of convergence occurs at an adversarial regret of $224$. As for PPO2, this value is at $268.2$. These results indicate a similar ranking of the robustness in these policies, with DQN being the least-robust to maximum of 5 perturbations, and the A2C prevailing as the most robust policy to maximum of 5 perturbations.

\subsection{Case 1: $\delta_{max} = 10$:}

\subsection{Case 2: $\delta_{max} = 5$:}

\section{Conclusion}
\label{sec:Conclusion}
We presented two RL-based techniques for benchmarking the resilience and robustness of DRL policies to adversarial perturbations of state transition dynamics. Experimental evaluation of our proposals demonstrate the feasibility of these techniques for quantitative analysis of policies with regards to their sensitivity to state transition dynamics. A promising venue of further exploration is to study and extend the proposed methodologies for evaluation of generalization in DRL policies. 

%
%
%
 \bibliographystyle{splncs04}
 \bibliography{ref}
%




\end{document}